\documentclass{esannV2}
\usepackage{enumitem}
\usepackage{graphicx}  % TODO: change this back!!
\usepackage[latin1]{inputenc}
\usepackage{amssymb,amsmath,array}
\usepackage{hyperref}
\usepackage{subcaption}
\usepackage{xspace}

\usepackage{todonotes}

\newcommand{\fodot}{FO($\cdot$)\xspace}
\newcommand{\figref}[1]{Fig.~\ref{#1}}

\begin{document}
%style file for ESANN manuscripts
\title{Enhancing Computer Vision with Knowledge:\\ a Rummikub Case Study}

%***********************************************************************
% AUTHORS INFORMATION AREA
%***********************************************************************
\author{Simon Vandevelde$^{1,2,3}$, Laurent Mertens$^{1,2}$, Sverre Lauwers$^1$\\ and Joost Vennekens$^{1,2,3}$
%
% Optional short acknowledgment: remove next line if non-needed
\thanks{This research received funding from the Flemish Government under the ``Onderzoeksprogramma Artifici\"ele Intelligentie (AI) Vlaanderen'' programme}
%
% DO NOT MODIFY THE FOLLOWING '\vspace' ARGUMENT
\vspace{.3cm}\\
%
% Addresses and institutions (remove "1- " in case of a single institution)
1- KU Leuven - Dept. Of Computer Science \\
J.-P. De Nayerlaan 5, 2860 Sint-Katelijne-Waver - Belgium
%
% Remove the next three lines in case of a single institution
\vspace{.1cm}\\
2- Leuven.AI - KU Leuven institute for AI \\
B-3000, Leuven - Belgium
\vspace{.1cm}\\
3- Flanders Make - DTAI-FET\\
}

%***********************************************************************
% END OF AUTHORS INFORMATION AREA
%***********************************************************************

\maketitle

\begin{abstract}
% Type your 100 words abstract here. 
% ANNs zijn heel goed in het identificeren van individuele componenten van een afbeelding. Het samenvoegen van die componenten tot een correcte interpretatie van een volledige scene blijkt echter heel wat moeilijker. Zoals te lezen in een prachtige recente NeurIPS paper, is dit bijvoorbeeld het geval als je de sociale context van een afbeelding probeert te interpreteren. In deze paper laten we zien dat je zelfs voor een veel eenvoudiger probleem al merkt dat individuele componenten correct geinterpreteerd worden, maar daarom nog niet de volledige afbeelding.
% Een mogelijke manier om dat te verbeteren is door expliciete achtergrond toe te voegen over het verband tussen de verschillende componenten. In deze paper onderzoeken we hoeveel dergelijke achtergrondkennis "waard is", en we laten zien dat dit voor ons voorbeeld evenveel waard is als twee derde van de dataset.
% \end{abstract}
Artificial Neural Networks excel at identifying individual components in an image.
However, out-of-the-box, they do not manage to correctly integrate and interpret these components as a whole.
One way to alleviate this weakness is to expand the network with explicit knowledge and a separate reasoning component. In this paper, we evaluate an approach to this end, applied to the solving of the popular board game Rummikub. We demonstrate that, for this particular example, the added background knowledge is equally valuable as two-thirds of the data set, and allows to bring down the training time to half the original time.
\end{abstract}

% ############################################################
\section{Introduction}
Artificial Neural Networks (ANNs) are considered a tried and tested method to identify objects in an image.
However, correctly interpreting these objects in relation to each other to form a complete picture remains difficult, as demonstrated in the literature~\cite{Mertens2024,Nogiets}.
One way to overcome this issue is by adding explicit \emph{background knowledge} about the identified items and their relations.
In this work, we apply this approach to the popular boardgame Rummikub. Concretely, we consider the task of correctly detecting all the tiles in a photo of a Rummikub game state. We compare the performance of a ``vanilla'' ANN setup to one that extends this setup with explicit knowledge and reasoning by means of the IDP-Z3~\cite{IDPZ3} system. In the following paragraphs, we introduce Rummikub and IDP-Z3.

\begin{figure}[!h]
    \centering
    \begin{subfigure}{.49\linewidth}
        \centering
        \includegraphics[width=0.7\linewidth]{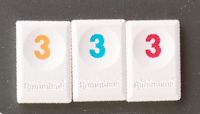}
        \caption{Example of a group.}
        \label{fig:group}
    \end{subfigure}
    \begin{subfigure}{.49\linewidth}
        \centering
        \includegraphics[width=\linewidth]{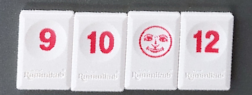}
        \caption{Example of a run, containing a joker.}
        \label{fig:run}
    \end{subfigure}
    \caption{Rummikub set examples}
    \label{fig:sets}
\end{figure}

% ------------------------------------------------------------
% Rummikub
\emph{Rummikub}\footnote{\url{https://rummikub.com}} is a popular board game in which players are given tiles defined by a number $n \in \left[1, 2,\dots, 13\right]$ and a color $c \in \left[\mbox{red}, \mbox{blue}, \mbox{black}, \mbox{yellow}\right]$. To win the game, players need to be the first to place all their tiles in the center of the game field. Tiles may only be played when they correctly form a \emph{set}. Two types of sets exist: a \emph{group}, in which 3 or 4 tiles share the same $n$ but have different $c$, and a \emph{run}, which is a series of 3 to 13 tiles of same $c$ with subsequent $n$. The game also contains two joker tiles which may be used as ``wildcards'' to form sets, as indicated by a smiling face. All these concepts are illustrated in \figref{fig:sets}.

% ------------------------------------------------------------
% IDP-Z3
\emph{IDP-Z3} is a logical reasoning engine for first order logic.
It adheres to the Knowledge Base Paradigm~\cite{KBP}, which states that knowledge should be modeled \textit{declaratively} in a Knowledge Base (KB) regardless of its use.
As such, the KB is merely a ``bag of knowledge''.%, which can not be executed like a normal program.
To put this knowledge to use, it can be given to a reasoning engine such as IDP-Z3, which supports several kinds of inference tasks.
This approach supports reusability: once formalized, the KB can be re-used to solve different types of problems in the same problem domain without modifications.
I.e., the same KB could be used to check satisfiability, find (optimal) solutions, derive consequences, explain incorrectness, etc.

The specific modeling language used for IDP-Z3 is \fodot, which extends classical first order logic by adding concepts (e.g., types and aggregates) to make modeling more user-friendly.
% We will briefly go over \fodot using the Rummikub rules as an example, and refer to~\cite{??} for a more complete overview.
As an example, consider the following logical formula which formalizes the rule of a correct Rummikub group.
\begin{equation*}
\begin{aligned}
\forall t_1, t_2 \in \mathit{Tile}: t_1 \neq t_2& \land set(t_1) = set(t_2) \land group(set(t_1)) \Rightarrow\\
& \mathit{number}(t_1) = \mathit{number}(t_2) \land \mathit{color}(t_1) \neq \mathit{color}(t_2).
\end{aligned}
\end{equation*}
%\blue{alternatief:
%\begin{equation*}
%\forall (t_1, t_2) \in \mathit{Group}: \mathit{number}(t_1) = \mathit{number}(t_2) \land \mathit{color}(t_1) \neq \mathit{color}(t_2).\\
%\end{equation*}
%}
This can be read as: ``for all different tiles $t_1, t_2$ within a group must hold that their numbers are the same and their colors are different''.\footnote{For a more detailed explanation on \fodot, we refer to~\cite{IDPZ3}.}

% \begin{equation*}
%     \begin{aligned}
%     & \forall t_1, t_2 \in \mathit{Tile}: \mathit{color}(t_1) = \mathit{color}(t_2).\\
%     & \forall t_1 \in \mathit{Tile}: \mathit{number}(t_1 - 1) = \mathit{number}(t_1) - 1 \lor t_1 = 0.
%     \end{aligned}
% \end{equation*}

% ############################################################
\section{Methodology}
We use a custom image dataset consisting of 285 manually captured images of Rummikub playing fields, employing three different zoom levels, four different lighting levels and two different backgrounds.
Special attention was paid to ensure the images are realistic: they all contain a varying number of only valid sets, at various positions and angles.
Each set ranges from 3 to 13 tiles, and is diverse in terms of colors and numbers. The tiles are also often not perfectly aligned, much like they would be in a real game.
Images were annotated for tile bounding boxes and tile number/color, with a total of 4336 tiles annotated.
Our dataset is publicly available through Kaggle~\footnote{\url{https://www.kaggle.com/datasets/sverrela/rummikub}}.

\begin{figure}
    \centering
    \includegraphics[width=\linewidth]{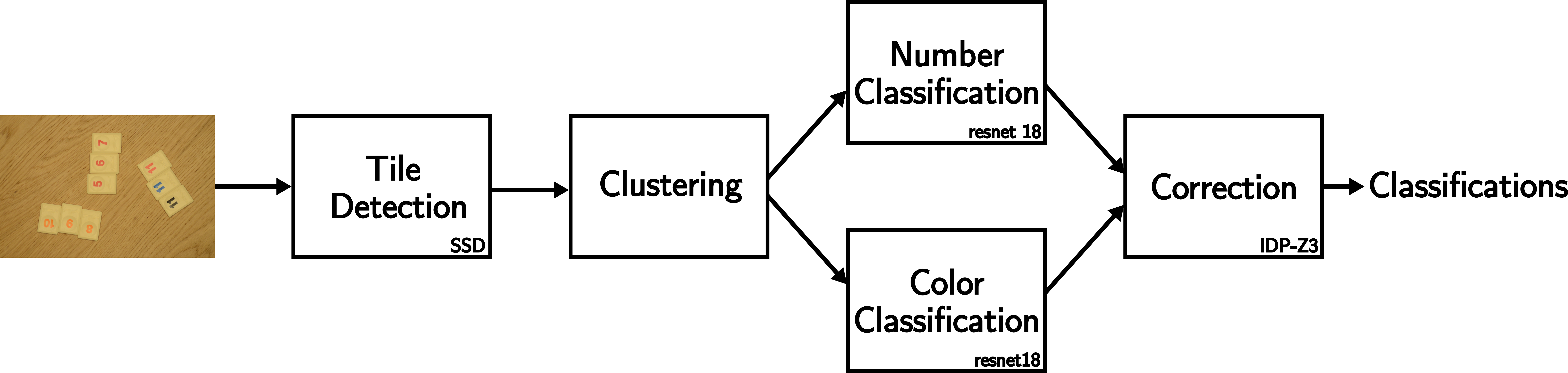}
    \caption{Overview of the tile detection/classification pipeline}
    \label{fig:pipeline}
\end{figure}

To correctly detect all tiles in a depicted Rummikub game state, we propose the pipeline shown in Fig.~\ref{fig:pipeline}. It consists of four steps:% first, we detect all tiles in the image; next, we cluster the tiles per set; then we detect the number and color of each tile; finally, logical reasoning is used to decide the most likely tile configuration. Crucially, we assume that the shown state is valid. In more detail:

\begin{enumerate}[label={\emph{Step \arabic*} -}, itemindent=1\parindent, leftmargin=2\parindent, itemsep=0pt]
\item \emph{Tile detection}:
Bounding boxes are generated for each individual tile in the image by means of a standard single shot multi-box detector (SSD)~\cite{SSD} trained on our dataset.
\item \emph{Clustering}:
A hand-crafted algorithm clusters the individual bounding boxes into sets.
This clustering algorithm is designed to be robust to size and orientation, so that it can handle the random nature of the sets.
\item \emph{Number/Color classification}:
Each detected tile is classified for $n$ and $c$ by means of two ResNet18~\cite{resnet} networks trained on our dataset.
The output of each network represents the confidence levels assigned to each possible $n$ and $c$ for each tile.
A pure ANN setup would at this point assign the class with the highest confidence to each tile.
Instead, we pass all confidence values on to the next step for further processing.
\item \emph{Correction}:
In this final step, the confidence levels obtained in Step~3 are used to model an optimization problem at the level of an entire set. Concretely, instead of assuming the most likely class for each tile individually, we try to find the most likely class for \textit{all} tiles in a set, given that the set must be correct (i.e., we explicitely assume the shown game state to be valid).
This way, we are able to enrich our detections with background knowledge.
\end{enumerate}

The optimization problem in Step 4 is straightforward.
We model the confidences in IDP-Z3 using binary functions of the form $\mathit{attr\_conf}: Tile \times Attr \rightarrow Number$, and add them to our Rummikub KB.
To compute the overall confidence score of a possible solution, we use the following straightforward sum:
\begin{equation*}
    \begin{aligned}
        \mathit{acc}() = & \sum_{t \in \mathit{Tile}, c \in \mathit{Color}} \mathit{color\_conf}(t, c) | \mathit{color}(t) = c \\
        + & \sum_{t \in \mathit{Tile}, n \in \mathit{Number}} \mathit{number\_conf}(t, n) | \mathit{number}(t) = n.
    \end{aligned}
\end{equation*}
In other words, we sum the confidences of the colors/numbers that have been assigned to the tiles via the $\mathit{color}$ and $\mathit{number}$ functions.
We then let IDP-Z3 find the optimal assignments of these functions, giving us the classifications with the highest confidences that are feasible.

%To clarify this with a concrete example, consider three tiles with the following color confidences:
As a concrete example, consider the following color confidences :
\begin{equation*}
    \begin{aligned}
        \mbox{Tile 1: } & (1, red) \rightarrow 0.8, (1, blue) \rightarrow 0.1, (1, black) \rightarrow 0.05, (1, orange) \rightarrow 0.05,\\
        \mbox{Tile 2: } & (2, red) \rightarrow 0.2, (2, blue) \rightarrow 0.7, (2, black) \rightarrow 0.09, (2, orange) \rightarrow 0.01,\\
        \mbox{Tile 3: } & (3, red) \rightarrow 0.5, (2, blue) \rightarrow 0.15, (2, black) \rightarrow 0.05, (2, orange) \rightarrow 0.3.\\
    \end{aligned}
\end{equation*}
% \begin{equation*}
%     \begin{aligned}
%         \mathit{color\_conf} & := \{\\
%         & (1, red) \rightarrow 0.8, (1, blue) \rightarrow 0.1, (1, black) \rightarrow 0.05, (1, orange) \rightarrow 0.05\\
%                                & (2, red) \rightarrow 0.2, (2, blue) \rightarrow 0.7, (2, black) \rightarrow 0.09, (2, orange) \rightarrow 0.01\\
%                                & (3, red) \rightarrow 0.5, (2, blue) \rightarrow 0.15, (2, black) \rightarrow 0.05, (2, orange) \rightarrow 0.3\\
%         & \}.
%     \end{aligned}
% \end{equation*}
Ignoring the numbers for the sake of the example, and assuming the set is a group, all tiles must therefore have a different color.
As such, instead of the highest likely \emph{individual} colors \emph{(red, blue, red)}, IDP-Z3 will correct these classifications to \emph{(red, blue, orange)} as the most confident \emph{feasible} classification.

% ############################################################
\section{Results}
All code required to reproduce the results discussed in this section is available from our dedicated GitLab repository\footnote{\url{https://gitlab.com/EAVISE/sva/knowledge-enhanced-rummikub-detector}}. All evaluations are performed using an Intel Xeon E5-2630 v3 with an NVIDIA Quadro P2000 and 32 GB of memory, using our full dataset as test data.\footnote{During training, the ResNet18 networks are being fed the annotated bounding boxes, while during evaluation, they receive their input from the SSD network, which is slightly different. Note that, while typically using the same data for train and testing purposes is considered bad practice, we argue that in this particular case it allows to evaluate the pure ANN approach in ``ideal'' circumstances, further highlighting the added benefit of adding a logical reasoning step.} We ran each experiment 10 times, and report averages and standard deviations.

\begin{figure}
    \centering
    \begin{subfigure}{0.49\linewidth}
        \includegraphics[width=\linewidth]{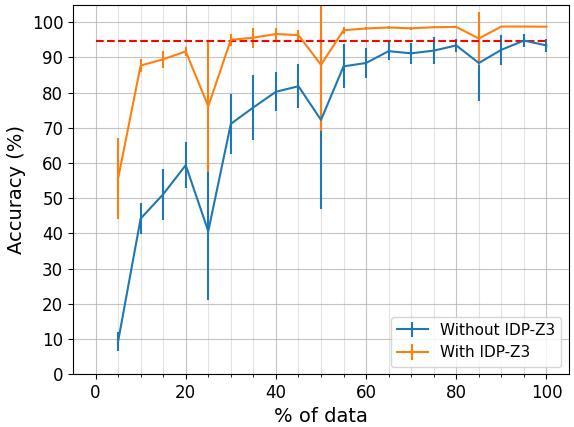}
        \caption{Accuracy when training on \% of data}
        \label{fig:result_data}
    \end{subfigure}
    \begin{subfigure}{0.49\linewidth}
        \includegraphics[width=\linewidth]{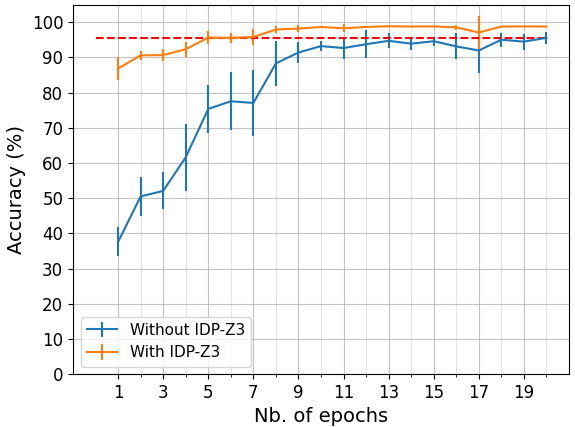}
        \caption{Accuracy when training for $x$ epochs}
        \label{fig:result_epoch}
    \end{subfigure}
    \caption{Ablation experiment results. The dashed red lines indicate the highest accuracy reached by the ANN-only approach.}
    \label{fig:results}
\end{figure}

%As a baseline, without using the IDP-Z3 step (Step 4), our pipeline reached %an individual tile accuracy of 99.49\% and
%an \emph{image accuracy}, defined as the percentage of images for which all tiles were correctly classified, of 93.40$\pm$1.96\%. By contrast, when adding the IDP-Z3 step, %single tile accuracy increased to 99.66\% (+0.17) and image accuracy increased to 98.73$\pm$0.18\% (+5.33\% on average).
%Recall that the accuracy of an ANN typically scales logarithmically w.r.t.\ training time and data, making the last percentage increases the most difficult to obtain.
We evaluated the pipeline in terms of the size of the training data, by limiting the available data to a \% subset, as shown in Fig~\ref{fig:result_data}.
For every run, ResNet18 color and number models were trained for 5 and 20 epochs respectively.
A few observations are noteworthy.
% To get a better view on the effect of the knowledge when not much data is present, Fig.~\ref{fig:result_data} shows the accuracies when only training the classifiers on a \% subset of the data.
%To this end, we trained and evaluated the classifiers multiple times, each time on a \% subset of the data, as shown in Fig~\ref{fig:result_data}.
%This shows a few interesting results.
% Firstly, as expected, the classifiers do not reach a high accuracy at only 5\% of the data (217 images).
First, when using only 5\% of the data, the context-based correction step greatly increases the total image accuracy from 9.31$\pm$2.63\% to 55.56$\pm$11.50\%.
%Second, the accuracy of the complete pipeline seems to ``follow'' the accuracy of the classifiers, with the percentage difference decreasing as the amount of data increases.
Second, the full pipeline outperforms the pure ANN approach, even at its highest reached accuracy (98.76$\pm$0.15\% @ 90\% vs. 94.79$\pm$1.85\% @ 95\% resp., or a 3.97\% increase).
% Third, and most interesting, the highest accuracy reached by the classifiers is already reached when using only 40\% of the data when taking detection context into account, as indicated by the dashed red line.
Third, it seems that, in general, adding the reasoning step seems to lower the standard deviations, acting like a ``stabiliser'' of sorts.
For example, between 30 and 45\% of the data, the ANNs had a standard deviation between 5.57 and 8.5, while the full pipeline's standard deviation remained between 1.37 and 2.8.
Fourth, and most interesting, when taking the detections into account, the highest accuracy reached by the classifiers is already reached when using only about 30\% of the data (with IDP-Z3 @ 30\%: 94.98$\pm$1.79\%, without IDP-Z3 @ 95\%: 94.79$\pm$1.85\%), as indicated by the dashed red line.

To evaluate the effect of training time, Fig.~\ref{fig:result_epoch} shows an analogous experiment conditioning on the number of epochs instead of training data. Here, both the ResNet18 color and number models are trained for $x$ epochs.
The same tendencies are apparent, though the initial jump in accuracy seems to be notably higher (37.61$\pm$4.14\% to 86.83$\pm$3.32\% after one epoch).
Again, the highest accuracy reached by the classifiers is achieved much earlier when adding logical reasoning: 5 epochs are sufficient instead of 20 (with IDP-Z3 @ 5 epochs: 95.64$\pm$1.84\%, without IDP-Z3 @ 20 epochs: 95.56$\pm$1.66\%).
Similarly, the full pipeline outperforms the pure ANN approach (98.84$\pm$0.00\% @ 13 epochs vs. 95.56$\pm$1.66\% @ 20 epochs resp.), and the standard deviations have decreased across the board.

It is, however, important to note that the correction step does add additional inference time, albeit the effect appears to be rather limited.
In our experiments, tile detection, clustering, and classifying required about 0.04s per tile, while correcting an entire set requires another 0.095s.
E.g., for a set consisting of 5 tiles the pipeline would require 0.2s to generate the classifications, and another 0.1s to correct them if necessary.

% \simon{Ook vermelden wanneer de pipleine niet werket -- i.e., wanneer we clustering problemen hadden} 

% ############################################################
\section{Related Work}
In recent years, there has been a rise in approaches aimed at extending ANNs with explicit reasoning.
The most notable approach, as shown in works such as DeepProbLog~\cite{Manhaeve2018} and NeurASP~\cite{Yang_Ishay_Lee_2020}, combine the two on a true neurosymbolic level, tightly coupling the neural networks with logic.
In this approach, the logic program is evaluated throughout the training process itself, resulting in increasingly higher accuracies.
However, in general, such systems have difficulties with scaling to complex domains, and are slow to train.
For a more complete overview of neurosymbolic approaches, we refer to~\cite{Marra2024}.
% For a more extensive overview of hybrid AI techniques, we refer to~\cite{Cornelio2023}

The work that comes closest to ours is that of Mulamba et al.~\cite{Mulamba2020}, who present a similar pipeline approach, but for the detection of digits in a (partial) Sudoku.
However, they did not evaluate the effect of data or training time, and instead only report on the final accuracy. Furthermore, the unknown tile positions in Rummikub, as opposed to fixed grid positions for Sudoku, make it a harder problem to solve.
% Werk v Tias - maar Rummikub is moeilijker doordat de posities onbekend zijn, en  wij checken meer het effect v.d. knowledge 

% ############################################################
\section{Conclusion}
We have demonstrated an approach to improve the accuracy of a computer vision system for the detection and validation of Rummikub game states, by adding explicit reasoning on background knowledge.
Through our evaluations, we have shown that, for this problem, background knowledge is worth as much as two-thirds of the data set, or slightly more than half of the training time.
In this sense, our approach is most useful in situations where data is scarce or difficult to gather, or when the ANNs are constraint by hardware limitations, such as in edge devices.
However, our approach can only succeed when there exists a clear relationship between all output classes, which is not always the case.
Among others, examples of real-life applications satisfying this constraint include sensor fusion and input detection in forms (e.g., tax forms).
% Handig wanneer (a) er duidelijke kennis aanwezig is (sensor fusion?), (b) er niet veel data om handen is of het NN niet performant is (b.v. edge).

As part of future work, we intend to compare our approach to a more neurosymbolic approach by implementing the Rummikub example in, e.g., NeurASP. We also plan to extend the work to allow the pipeline to suggest corrections when errors (i.e., invalid sets) are present in the image. Finally, we expect to further evaluate our pipeline on some of the real-life problem domains mentioned above.

\begin{footnotesize}

\bibliographystyle{unsrt}
\bibliography{biblio}

\begin{thebibliography}{10}

\bibitem{Mertens2024}
Laurent Mertens, Elahe' Yargholi, Hans~Op {de Beeck}, Jan~Van {den Stock}, and Joost Vennekens.
\newblock {{FindingEmo}}: {{An}} image dataset for emotion recognition in the wild (accepted at {NeurIPS} 2024).
\newblock 2024.

\bibitem{Nogiets}
Oge Marques, Elan Barenholtz, and Vincent Charvillat.
\newblock Context modeling in computer vision: Techniques, implications, and applications.
\newblock {\em Multimedia Tools and Applications}, 51(1):303--339, January 2011.

\bibitem{IDPZ3}
Pierre Carbonnelle, Simon Vandevelde, Joost Vennekens, and Marc Denecker.
\newblock {{IDP-Z3}}: A reasoning engine for {{FO}}(.).
\newblock {\em arXiv preprint arXiv:2202.00343}, 2022.

\bibitem{KBP}
Marc Denecker and Joost Vennekens.
\newblock Building a {{Knowledge Base System}} for an {{Integration}} of {{Logic Programming}} and {{Classical Logic}}.
\newblock In {\em Logic {{Programming}}}, volume 5366, pages 71--76. Springer Berlin Heidelberg, Berlin, Heidelberg, 2008.

\bibitem{SSD}
Wei Liu, Dragomir Anguelov, Dumitru Erhan, Christian Szegedy, Scott Reed, Cheng-Yang Fu, and Alexander~C. Berg.
\newblock Ssd: Single shot multibox detector.
\newblock In {\em Computer Vision -- ECCV 2016}, pages 21--37, Cham, 2016. Springer International Publishing.

\bibitem{resnet}
Kaiming He, Xiangyu Zhang, Shaoqing Ren, and Jian Sun.
\newblock Deep residual learning for image recognition.
\newblock In {\em Proceedings of the IEEE Conference on Computer Vision and Pattern Recognition (CVPR)}, June 2016.

\bibitem{Manhaeve2018}
Robin Manhaeve, Sebastijan Dumancic, Angelika Kimmig, Thomas Demeester, and Luc De~Raedt.
\newblock Deepproblog: {{Neural}} probabilistic logic programming.
\newblock {\em Advances in neural information processing systems}, 31, 2018.

\bibitem{Yang_Ishay_Lee_2020}
Zhun Yang, Adam Ishay, and Joohyung Lee.
\newblock Neurasp: Embracing neural networks into answer set programming.
\newblock In {\em Proceedings of IJCAI-20}, pages 1755--1762. International Joint Conferences on Artificial Intelligence Organization, July 2020.

\bibitem{Marra2024}
Giuseppe Marra, Sebastijan Duman{\v c}i{\'c}, Robin Manhaeve, and Luc De~Raedt.
\newblock From statistical relational to neurosymbolic artificial intelligence: {{A}} survey.
\newblock {\em Artificial Intelligence}, 328:104062, March 2024.

\bibitem{Mulamba2020}
Maxime Mulamba, Jayanta Mandi, Rocsildes Canoy, and Tias Guns.
\newblock Hybrid {{Classification}} and {{Reasoning}} for {{Image-Based Constraint Solving}}.
\newblock In {\em Integration of {{Constraint Programming}}, {{Artificial Intelligence}}, and {{Operations Research}}}, pages 364--380, 2020.

\end{thebibliography}

\end{footnotesize}

% ****************************************************************************
% END OF BIBLIOGRAPHY AREA
% ****************************************************************************

\end{document}